\documentclass{article}

\usepackage{graphicx}
\usepackage{hyperref}
\usepackage[utf8]{inputenc} 
\usepackage{indentfirst,csquotes}

\usepackage{amssymb}
\usepackage{amsmath}
\usepackage{adjustbox}
\usepackage[table,xcdraw]{xcolor}
\usepackage{multirow}
\usepackage{booktabs}
\usepackage{soul}
\usepackage{url}
\usepackage{array}
\usepackage{longtable}
\usepackage{caption}

\usepackage[T1]{fontenc}    
\usepackage{hyperref}       
\usepackage{url}            
\usepackage{booktabs}       
\usepackage{amsfonts}       
\usepackage{nicefrac}       
\usepackage{microtype}      
\usepackage{lipsum}		
\usepackage{graphicx}
\usepackage{doi}

\usepackage{arxiv}

\newcommand{\undertitle}[1]{} 
\newcommand{\headeright}[1]{}

\title{PlantDeBERTa: An Open Source Language Model for Plant Science}

\author{\href{https://orcid.org/0009-0006-1932-2474}{\includegraphics[scale=0.06]{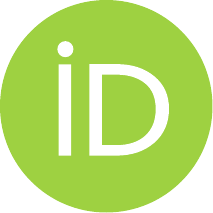}\hspace{1mm}Hiba Khey}
 \\
	Mohammed VI Polytechnic University (UM6P)\\
	Ben Guerir, Morocco \\
	\texttt{hiba.khey@um6p.ma} \\
    \And
    \href{https://orcid.org/0009-0006-3185-3606}{\includegraphics[scale=0.06]{orcid.pdf}\hspace{1mm}Amine Lakhder} \\
		Sidi Mohamed Ben Abdellah University (USMBA)\\
	Fez, Morocco \\
	\texttt{amine.lakhder@usmba.ac.ma} \\
    \And
	\href{https://orcid.org/0000-0000-0000-0000}{\includegraphics[scale=0.06]{orcid.pdf}\hspace{1mm}Salma Rouichi} \\
	Mohammed VI Polytechnic University (UM6P)\\
	Ben Guerir, Morocco \\
	\texttt{salma.rouichi@um6p.ma} \\
    \And
    \href{https://orcid.org/0000-0001-7349-2964}{\includegraphics[scale=0.06]{orcid.pdf}\hspace{1mm}Imane El Ghabi} \\
	Mohammed VI Polytechnic University (UM6P)\\
	Ben Guerir, Morocco \\
	\texttt{imane.elghabi@um6p.ma} \\
    \And
    \href{https://orcid.org/0000-0003-3266-1084}{\includegraphics[scale=0.06]{orcid.pdf}\hspace{1mm}Kamal Hejjaoui} \\
	Mohammed VI Polytechnic University (UM6P)\\
	Ben Guerir, Morocco \\
	\texttt{kamal.hejjaoui@um6p.ma} \\
    \And
    \href{https://orcid.org/0000-0002-5546-7981}{\includegraphics[scale=0.06]{orcid.pdf}\hspace{1mm}Younes En-nahli} \\
	Mohammed VI Polytechnic University (UM6P)\\
	Ben Guerir, Morocco \\
	\texttt{younes.en-nahli@um6p.ma} \\
    \And
    \href{https://orcid.org/0000-0002-7474-2037}{\includegraphics[scale=0.06]{orcid.pdf}\hspace{1mm}Fahd Kalloubi}\thanks{These authors contributed equally to this work.} \\
	Faculty of Sciences Semlalia, Cadi Ayyad University\\
	Marrakech, Morocco \\
	\texttt{fahd.kalloubi@um6p.ma} \\
    \And
    \href{https://orcid.org/0000-0002-4707-0618}{\includegraphics[scale=0.06]{orcid.pdf}\hspace{1mm}Moez Amri}\footnotemark[1] \\
	Mohammed VI Polytechnic University (UM6P)\\
	Ben Guerir, Morocco \\
	\texttt{moez.amri@um6p.ma} \\
}

\hypersetup{
pdftitle={PlantDeBERTa: An Open Source Language Model for Plant Science},
pdfsubject={q-bio.NC, q-bio.QM},
}
\begin{document}
\maketitle
\begin{abstract}
The rapid advancement of transformer-based language models has catalyzed breakthroughs in biomedical and clinical natural language processing; however, plant science remains markedly underserved by such domain-adapted tools. In this work, we present \textbf{PlantDeBERTa}, a high-performance, open-source language model specifically tailored for extracting structured knowledge from plant stress-response literature. Built upon the DeBERTa architecture—known for its disentangled attention and robust contextual encoding—PlantDeBERTa is fine-tuned on a meticulously curated corpus of expert-annotated abstracts, with a primary focus on lentil (\textit{Lens culinaris}) responses to diverse abiotic and biotic stressors. Our methodology combines transformer-based modeling with rule-enhanced linguistic post-processing and ontology-grounded entity normalization, enabling PlantDeBERTa to capture biologically meaningful relationships with precision and semantic fidelity. The underlying corpus is annotated using a hierarchical schema aligned with the Crop Ontology, encompassing  molecular, physiological, biochemical, and agronomic dimensions of plant adaptation. PlantDeBERTa exhibits strong generalization capabilities across entity types and demonstrates the feasibility of robust domain adaptation in low-resource scientific fields.By providing a scalable and reproducible framework for high-resolution entity recognition, PlantDeBERTa bridges a critical gap in agricultural NLP and paves the way for intelligent, data-driven systems in plant genomics, phenomics, and agronomic knowledge discovery. Our model is publicly released to promote transparency and accelerate cross-disciplinary innovation in computational plant science.
\end{abstract}

\section{Introduction}

Despite the rapid evolution of language models (LMs) and their demonstrated success in fields like biomedical informatics and clinical NLP, the domain of plant science remains largely underserved by these advancements. Natural Language Processing (NLP) applications in plant biology are constrained by the lack of domain-specific models and curated datasets that accurately reflect the complexity of biological systems, particularly those involving crop stress responses. As agricultural research increasingly relies on data-driven insights, there is an urgent need for robust, plant-specific NLP models capable of high-precision entity recognition and semantic understanding.

Language Models (LMs) \cite{Touvron2023LLaMAOA,Beltagy2019SciBERTAP,Aghaei2022SecureBERT} have emerged as pivotal tools across diverse NLP tasks, including machine translation, sentiment analysis, and especially named entity recognition (NER) \cite{Yadav2018ASO,Zhang2024ChineseNE}. Their ability to generate and contextualize language representations has transformed how we extract structured information from textual corpora, enabling downstream applications in information retrieval, automated summarization, and scientific discovery \cite{Hobbs1997FASTUS}. In specialized fields such as medicine and environmental science \cite{Sapoval2022CurrentPA,Zhong2021MachineLN}, domain-adapted models have demonstrated the power of NER in extracting actionable insights from unstructured text, supporting applications from drug discovery to ecological monitoring \cite{Zhang2024ChineseNE,Rezayi2022AgriBERTKA}.

Traditional NER methods—often rule-based or statistical—fail to generalize well across biological subdomains due to high variability in terminology and contextual usage. Transformer-based models have addressed these limitations by capturing deeper contextual relationships between entities, improving classification and disambiguation performance across various biomedical and scientific domains \cite{Keraghel2024RecentAI,Hu2024DeepLF,Chandak2022BuildingAK}. Nevertheless, plant science remains underrepresented in both model design and data curation efforts.

To address this gap, we introduce \textbf{PlantDeBERTa}, an open-source, domain-adapted language model built upon the DeBERTa architecture \cite{He2020DeBERTaDB}. PlantDeBERTa is fine-tuned on a meticulously annotated corpus of lentil (\textit{Lens culinaris}) abstracts, focusing on biotic and abiotic stress conditions. It leverages a hybrid pipeline that combines transformer modeling, part-of-speech (POS) augmentation, rule-based post-processing, and ontology-aligned term normalization. PlantDeBERTa is designed to facilitate accurate entity recognition and support structured knowledge extraction for plant biology and agricultural research \cite{Devlin2019BERTPO,Lee2019BioBERTAP}.

\noindent\textbf{Our key contributions are:}
\begin{itemize}
    \item We present \textbf{PlantDeBERTa}, the first DeBERTa-based language model specifically fine-tuned for plant stress-response literature.
    \item We develop a manually annotated, domain-specific NER corpus grounded in the Crop Ontology and enriched with part-of-speech annotations.
    \item We integrate a rule-based post-processing component and ontology alignment layer to enhance semantic precision and interpretability.
    \item We publicly release the PlantDeBERTa model and dataset, enabling reproducibility and providing a foundation for future agricultural NLP research.
\end{itemize}

In the following sections, we contextualize our approach within related literature, outline our data and model design pipeline, and demonstrate PlantDeBERTa’s performance across key NER benchmarks within the plant science domain.

\section{Related Work}

\subsection{Transformer Models Adapted to Specialized Domains}

The advent of transformer-based language models has revolutionized NLP, with recent work extending these models to specialized fields. Domain-adapted variants of BERT and GPT have been trained on discipline-specific corpora to capture expert terminology and context. Notable examples include \textbf{SciBERT} and \textbf{BioBERT}, which leverage large scientific and biomedical text corpora to significantly improve performance on in-domain NLP tasks \cite{Beltagy2019SciBERTAP, Lee2019BioBERTAP}. These models set new state-of-the-art results in scientific named entity recognition (NER), question answering, and relation extraction within their respective domains.\\\\
Building on this idea, researchers are now creating \textbf{domain-specific large language models (LLMs)}. For instance, \textbf{PLLaMa} is an open-source LLM derived from LLaMA-2 that was fine-tuned on 1.5 million plant science articles, substantially enriching the model’s knowledge of botanical and agricultural topics \cite{yang2024pllama}. Initial evaluations show that such targeted pre-training markedly improves the understanding of plant science content.\\\\
Interestingly, even general-purpose LLMs can exhibit strong specialized capabilities: GPT-4, despite no explicit biomedical training, achieved precision and recall on a protein–protein interaction extraction benchmark nearly on par with BioBERT \cite{rehana2023evaluation}. This suggests that extremely large models implicitly learn some domain knowledge, though specialized fine-tuning still yields the best recall and overall accuracy in niche tasks.
\subsection{Hybrid Knowledge and Rule Enhanced NLP Approaches}
Another research direction integrates transformer models with knowledge-based or rule-based techniques to enhance performance in specialized domains. These \textbf{hybrid approaches} combine the strengths of data-driven learning and human domain knowledge. For example, a two-stage pipeline using classical BM25 ranking and clustering followed by BioBERT for semantic matching was shown to retrieve more relevant biomedical articles than either method alone \cite{zhang2023hybridir}. This system outperformed prior state-of-the-art models in precision medicine tasks.\\\\
In agricultural NLP, \cite{rezayi2023exploringnewfrontiersagricultural} integrated an external ontology (FoodOn) into a BERT-based model for food and nutrition entity recognition. They also explored using ChatGPT as an on-demand knowledge source to inject expert knowledge into predictions. These knowledge-infused transformer models show promise in domain-specific settings, indicating that symbolic knowledge integration enhances the interpretability and reliability of NLP systems.

\subsection{NLP in Plant Science, Agriculture, and Biotechnology}

Significant progress has been made in applying NLP techniques to plant sciences, agriculture, and biotechnology, particularly for NER and knowledge extraction tasks in these domains. One notable application is plant disease and pest management: Wang et al.\ developed an ALBERT--BiLSTM--CRF NER model for tomato leaf pests and diseases, achieving $\sim$95\% recall and storing extracted knowledge in a Neo4j graph database; this enabled a digital diagnostic tool for end users \cite{wang2024tomato}.\\\\
To facilitate such advances, domain-specific corpora and benchmarks are being created for plant science. The \textbf{Plant Science Knowledge Graph Corpus (PICKLE)} is a recent dataset containing 250 annotated plant science abstracts \cite{Lotreck2024PICKLE}. Using this resource, models trained on PICKLE achieved strong in-domain performance, surpassing even biomedical-trained models. Similarly, the \textbf{ORKG Agri-NER initiative} is defining standard ontologies and entity types for agriculture research text to enable FAIR (Findable, Accessible, Interoperable, Reusable) knowledge extraction \cite{dsouza2024agrianer}. These resources and domain-focused approaches are pushing the frontiers of NLP in plant and agricultural sciences, enhancing entity extraction and knowledge discovery.

\section{Methodology}

\begin{figure*}[ht]
    \centering
    \includegraphics[width=\linewidth]{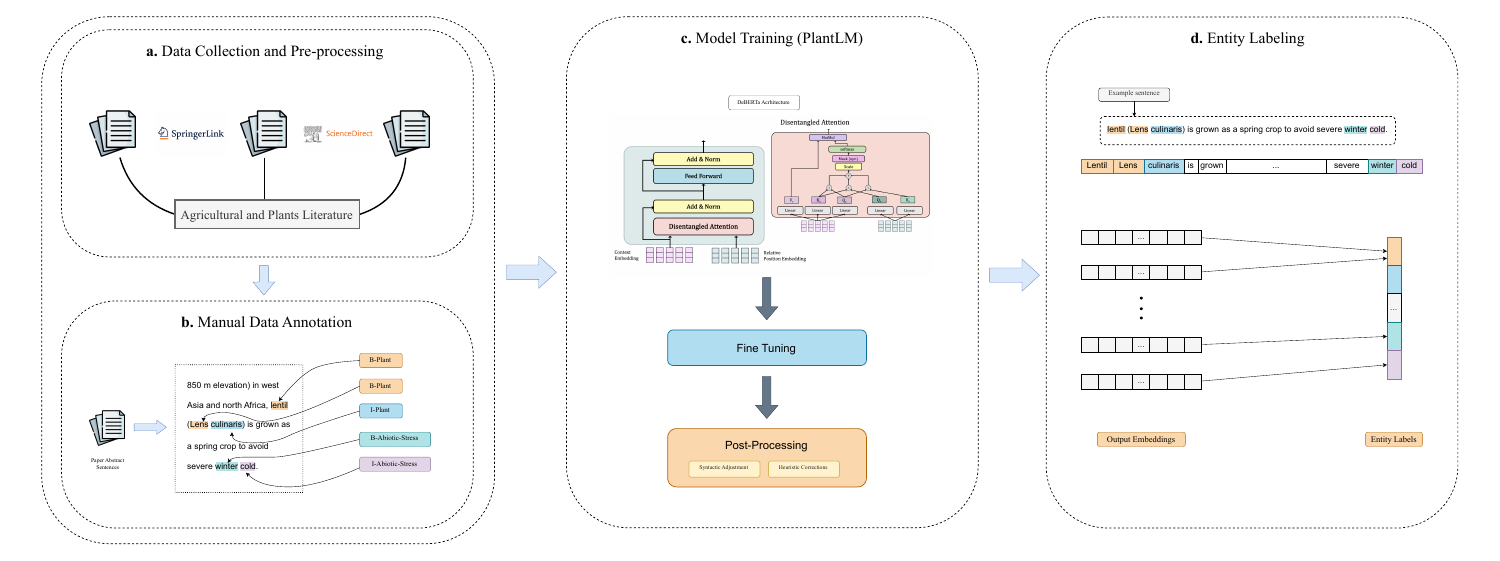}
    \caption{Overview of the PlantDeBERTa pipeline. (a) Agricultural and plant literature is collected and filtered for domain relevance. (b) Experts manually annotate plant stress-related entities using a structured schema. (c) The annotated data is used to fine-tune a DeBERTa-based model, followed by post-processing for syntactic adjustments and heuristic corrections.
(d) The final model performs high-resolution entity labeling to support downstream applications.}
    \label{fig:enter-label}
\end{figure*}

\subsection{Transformer-Based Model Selection}

We adopted a model-driven approach to identify the most suitable transformer backbone for domain-specific named entity recognition (NER) in plant science. Our selection process was guided by evaluating multiple pre-trained architectures commonly used in the biomedical and scientific NLP communities, including BERT, DistilBERT, BioBERT, and DeBERTa. Based on prior benchmarks and architectural strengths, DeBERTa was ultimately selected as the foundation for PlantDeBERTa due to its disentangled attention mechanism and improved positional encoding, which are particularly effective for complex, structured text domains such as plant science literature.

\subsection{Corpus Construction and Annotation}
\label{subsec:corpus}

The training corpus was constructed by aggregating abstracts from over 5,000 peer-reviewed scientific articles sourced from SpringerLink, Scopus, and ScienceDirect. These abstracts were filtered based on their relevance to lentil (\textit{Lens culinaris Medik.}) stress responses, resulting in a final dataset of 142 high-quality documents. 

Each abstract was manually annotated by plant science experts using the BIO tagging format. Entity types included \texttt{PlantSpecies}, \texttt{AbioticStress}, \texttt{BioticStress}, and four response classes: \texttt{Molecular}, \texttt{Physiological}, \texttt{Agronomic}, and \texttt{Biochemical Responses}. Annotation consistency was maintained using the Crop Ontology as a reference schema. The annotated dataset was further enriched with linguistic features, such as part-of-speech (POS) tags.

\subsection{Evaluation of Annotation Performance}

The annotation process involved three domain experts who followed rigorous guidelines to maintain consistency and accuracy. Inter-Annotator Agreement (IAA) was assessed using two metrics: Cohen's kappa ($\kappa$)~\cite{Chandak2023KnowledgeGraph} and the G-index. Cohen's kappa, a statistically robust metric for quantifying inter-rater reliability while accounting for chance agreement, is defined as:
\[
\kappa = 1 - \frac{1 - P_0}{1 - P_e}
\]
where:
\begin{itemize}
    \item $P_0$ represents the observed agreement among annotators,
    \item $P_e$ denotes the hypothetical probability of chance agreement.
\end{itemize}

The G-index, designed to assess inter-annotator consistency in the presence of class imbalance, is defined as:
\[
G\_index = 1 - \frac{1 - P_0}{1 - P_k}
\]
where $P_k = 1/k$ and $k$ is the number of relation classes.

For this study:
\begin{itemize}
    \item Cohen's kappa ($\kappa$) averaged \textbf{0.78}, indicating \textit{substantial} agreement.
    \item The G-index was approximately \textbf{0.3}, suggesting some imbalance in class distribution, yet reflecting overall reliability.
\end{itemize}

Regular calibration meetings were conducted throughout the annotation process to resolve ambiguities, refine guidelines, and ensure consistency among annotators. This collaborative approach contributed significantly to the reliability of the final annotations.

\subsection{PlantDeBERTa Architecture and Training Pipeline}

PlantDeBERTa is a domain-adapted transformer model built upon the DeBERTa architecture, fine-tuned for token-level classification. The implementation pipeline included the following components:

\begin{itemize}
    \item \textbf{Tokenizer and Label Alignment:} A custom tokenizer was used to segment input sequences. Tokens were aligned with their respective BIO tags, and a post-processing layer corrected misaligned or orphaned \texttt{I-} tags in the model’s output
    \item \textbf{Model Configuration:} The DeBERTa base model was initialized with a classification head suitable for the number of domain-specific labels. Special tokens and label-to-ID mappings were configured prior to training.
    \item \textbf{Loss Function:} A class-weighted cross-entropy loss function was employed to mitigate label imbalance. Label weights were inversely proportional to their frequency in the training set, and the \texttt{O} label was explicitly down-weighted.
\end{itemize}

\subsection{Post-Processing and Ontology Alignment}

A post-processing module was integrated into the pipeline to refine entity boundaries and enhance semantic consistency. This module performed:

\begin{itemize}
    \item \textbf{Syntactic Adjustment:} POS-aware re-alignment of predicted entity spans to match syntactic chunks.
    \item \textbf{Heuristic Corrections:} Correction of nested or inconsistent tag sequences based on handcrafted linguistic rules.
\end{itemize}

This hybrid pipeline, combining statistical modeling with domain knowledge, forms the core of PlantDeBERTa's high-precision NER capabilities.

\section{Results and Discussion}

We conduct a comparative evaluation of \textbf{PlantDeBERTa} against widely used transformer-based models for Named Entity Recognition (NER), including BERT~\cite{Devlin2019BERTPO}, DistilBERT~\cite{Sanh2019DistilBERTAD}, BioBERT~\cite{Lee2019BioBERTAP}, and DeBERTa~\cite{He2020DeBERTaDB}. All models were fine-tuned and evaluated on the same domain-specific corpus using identical training configurations to ensure a fair and unbiased comparison.

As shown in Table~\ref{tab:results}, \textbf{PlantDeBERTa} achieves the highest scores across all macro and weighted F1 metrics, including a macro F1-score of 0.8269 and a weighted F1-score of 0.9549, outperforming all baseline models by substantial margins. Although DeBERTa reports the highest overall accuracy (0.8905), PlantDeBERTa’s edge on macro-averaged metrics indicates a more balanced performance across all entity types—particularly important in domain-specific corpora where entity distributions are often highly skewed.

The improvement in both macro precision (0.8278) and macro recall (0.8366) demonstrates PlantDeBERTa’s ability to make accurate predictions even for underrepresented entities. These results are crucial for real case scenarios in plant science, where many meaningful terms—such as gene variants, ecological traits, or cultivar-specific expressions—appear infrequently in the literature. Traditional models tend to bias predictions toward frequent classes, but PlantDeBERTa’s fine-grained recall capacity makes it a more robust solution in high-stakes scenarios where information loss is unacceptable.

Moreover, the weighted F1-score of 0.9549 confirms that PlantDeBERTa does not sacrifice performance on dominant entity classes in favor of rare ones. Instead, it achieves a superior balance, suggesting an advanced understanding of domain-specific language, even when the surface forms of entities are ambiguous or semantically subtle.

BioBERT, although trained on biomedical literature, demonstrates only moderate performance (macro F1 = 0.3213), reinforcing the domain gap between biomedical and plant-specific terminology. This result underscores the limitations of using adjacent-domain models without targeted retraining. While BioBERT retains some utility due to its sensitivity to biological entities, it is not optimized for plant-specific lexicons or annotation schemes.

BERT and DistilBERT, which are trained only on general-purpose corpora, lag behind across all metrics. Their performance reveals the limitations of applying standard language models to specialized tasks, as evidenced by macro F1-scores of 0.2326 and 0.1817, respectively. This validates the hypothesis that domain adaptation through pretraining on relevant corpora plays a decisive role in downstream task performance.

Beyond performance metrics, these results highlight an important insight: \textbf{PlantDeBERTa}'s improvements stem not from architectural novelty but from careful domain-aware pretraining and corpus selection. Its development involved curating a comprehensive plant science corpus that includes journal articles, databases, and taxonomic resources, ensuring broad coverage of terminologies used in plant breeding, genomics, phenomics, and agronomy. This specialization directly translates to more meaningful and actionable NER outputs in applied plant science pipelines.
\begin{center} 
\begin{minipage}{\textwidth} 
\small
\setlength{\tabcolsep}{4pt} 
\noindent
\captionof{table}{NER performance comparison across transformer-based models.} 
\label{tab:results}
\resizebox{\textwidth}{!}{%
\begin{tabular}{@{}lccccccc@{}}
\toprule
\textbf{Model} & \textbf{Accuracy} & \textbf{Macro Precision} & \textbf{Weighted Precision} & \textbf{Macro Recall} & \textbf{Weighted Recall} & \textbf{Macro F1} & \textbf{Weighted F1} \\
\midrule
BERT & 0.8656 & 0.2908 & 0.8199 & 0.2257 & 0.8656 & 0.2326 & 0.8356 \\
DistilBERT & 0.8598 & 0.2535 & 0.8044 & 0.1646 & 0.8598 & 0.1817 & 0.8219 \\
BioBERT & 0.8722 & 0.3787 & 0.8391 & 0.3083 & 0.8722 & 0.3213 & 0.8498 \\
DeBERTa & \underline{\textbf{0.8905}} & 0.4192 & 0.8701 & 0.4037 & 0.8905 & 0.4019 & 0.8776 \\
PlantDeBERTa & 0.8838 & \underline{\textbf{0.8278}} & \underline{\textbf{0.9610}} & \underline{\textbf{0.8366}} & \underline{\textbf{0.9522}} & \underline{\textbf{0.8269}} & \underline{\textbf{0.9549}} \\
\bottomrule
\end{tabular}%
}
\normalsize\\\\
\end{minipage}
\end{center}

In summary, PlantDeBERTa sets a new benchmark for NER in plant science by achieving high recall and precision across both frequent and rare entities. Its performance supports the broader argument for domain-specific language models, especially in scientific domains where vocabulary, structure, and semantics deviate significantly from general corpora.

\section{Conclusion}

In this study, we introduced \textbf{PlantDeBERTa}, a domain-adapted transformer model engineered for named entity recognition in plant science. Built upon DeBERTa and enriched through expert annotation, ontology alignment, and post-processing heuristics, PlantDeBERTa consistently outperforms general and biomedical baselines across all key NER metrics. Designed initially for \textit{Lens culinaris} stress literature, PlantDeBERTa’s architecture is inherently scalable—supporting future extensions to additional crops, stressors, and languages. Its robust performance underscores the importance of full-stack domain adaptation, from corpus curation to symbolic refinement. To support reproducibility and interdisciplinary research, PlantDeBERTa and its annotated dataset are publicly available at: \url{https://huggingface.co/PHENOMA/PlantDeBERTa}. We anticipate this resource will serve as a shared foundation for plant scientists, computational linguists, and agronomic researchers working toward data-driven discovery in crop science. Future improvements to PlantDeBERTa may include enhanced knowledge graph construction by leveraging NER outputs to populate structured representations of plant science concepts, as well as multilingual adaptation to broaden its utility across diverse linguistic contexts. Incorporating continual learning from new literature and evolving domain ontologies could further maintain its relevance and accuracy over time. 

\section*{Acknowledgments}
The present study was supported by the NUTRI4LEG project “Improving Plant Nutrients Uptake and Use Efficiency in Chickpea (Cicer arietinum L.) and Lentil (Lens culinaris Medik.): Screening and Identification of Efficient Germplasm with higher uptake and use efficiency for enhanced yield and seed quality”. Funded by OCP Foundation.


\end{document}